\documentclass[runningheads]{llncs}
\usepackage{csquotes}
\usepackage{graphicx}
\usepackage{amssymb}
\usepackage{amsmath}
\usepackage{amsfonts}
\usepackage{multirow}
\usepackage[acronym]{glossaries}
\usepackage{hyperref}
\usepackage{mathtools}
\usepackage{dirtytalk}
\usepackage{caption}
\usepackage[english]{babel}

\begin{document}

\title{Time Matters: Time-Aware LSTMs for Predictive Business Process Monitoring}
\author{An Nguyen\inst{1}\thanks{Corresponding author}
\and
Srijeet Chatterjee\inst{1}
\thanks{Equal contribution with An Nguyen}
\and
Sven Weinzierl\inst{2}
\and
Leo Schwinn\inst{1}
\and
Martin Matzner\inst{2}
\and
Bjoern Eskofier\inst{1}
}
\authorrunning{A. Nguyen et al.}
\titlerunning{Time-Aware LSTMs for Predictive Business Process Monitoring}

\institute{Department of Computer Science, Friedrich-Alexander-University Erlangen-Nürnberg (FAU), Erlangen, Germany 
\email{\{an.nguyen, srijeet.chatterjee, leo.schwinn, bjoern.eskofier\} @fau.de}\\
\and
Insitute of Information Systems, Friedrich-Alexander-University Erlangen-Nürnberg (FAU), Nürnberg, Germany 
\email{\{sven.weinzierl, martin.matzner\} @fau.de}\\
 }
\maketitle 
\begin{abstract}
Predictive business process monitoring (PBPM) aims to predict future process behavior during ongoing process executions based on event log data. Especially, techniques for the next activity and timestamp prediction can help to improve the performance of operational business processes.
Recently, many PBPM solutions based on deep learning were proposed by researchers. Due to the sequential nature of event log data, a common choice is to apply recurrent neural networks with long short-term memory (LSTM) cells.
We argue, that the elapsed time between events is informative. However, current PBPM techniques mainly use \say{vanilla} LSTM cells and hand-crafted time-related control flow features.
To better model the time dependencies between events, we propose a new PBPM technique based on time-aware LSTM (T-LSTM) cells. T-LSTM cells incorporate the elapsed time between consecutive events inherently to adjust the cell memory.
Furthermore, we introduce cost-sensitive learning to account for the common class imbalance in event logs.
Our experiments on publicly available benchmark event logs indicate the effectiveness of the introduced techniques.

\keywords{Predictive Business Process Monitoring, Deep Learning, Recurrent Neural Network, LSTM, Time-Awareness.}
\end{abstract}
\newcommand{\magic}[0]{TLSTM} 

\newacronym{pbpm}{PBPM}{predictive business process monitoring}
\newacronym{bpm}{BPM}{business process management}
\newacronym{ml}{ML}{machine learning}
\newacronym{lstm}{LSTM}{long short-term memory}
\newacronym{dl}{DL}{deep learning}
\newacronym{dnn}{DNN}{deep neural network}
\newacronym{tlstm}{T-LSTM}{time-aware LSTM}
\newacronym{nn}{NN}{Neural Network}
\section{Introduction}

In the last years, a variety of \gls{pbpm} techniques that base on \gls{ml} were proposed by researchers~\cite{di.2018} to improve the performance of operational business processes~\cite{breuker2016comprehensible}. 
\gls{pbpm} is a class of techniques aiming at predicting future process characteristics in running process instances~\cite{maggi.2014}, like next activities, next timestamps or process-related performance indicators. Such \gls{pbpm} techniques produce predictions through predictive models. These models are in turn constructed based on historical event log data. 

A current trend in \gls{pbpm} is to apply \glspl{dnn} to learn more accurate predictive models from event log data than with \say{traditional} \gls{ml} algorithms like probabilistic automata~\cite{evermann2017predicting}. \glspl{dnn} belong to the \gls{ml}-sub-field \gls{dl} and achieve that by identifying the intricate structures in high-dimensional data through multi-representation learning~\cite{lecun.2015}.
\par
Existing \gls{dl}-based \gls{pbpm} techniques often rely on \gls{dnn} architectures consisting of out-of-the-box constructs like layers with a \textquote{vanilla} \gls{lstm} cell~\cite{hochreiter.1997} or state-of-the-art loss functions for parameter learning.

\par
 Event logs can be seen as sequences of events in continuous time with irregular intervals (i.e., elapsed time between consecutive events). We argue that these time intervals are informative in the case of event logs in \gls{pbpm}. Intuitively, these time intervals describe human behavior of executing business processes. Thus, a time-aware \gls{pbpm} technique considering information on time intervals could potentially achieve a higher predictive quality. Time information is currently only exploited via hand-crafted control-flow features as inputs to \textquote{vanilla} \gls{lstm} cells \cite{tax2017predictive}. To better account for the time information in event log data, we propose a new \gls{pbpm} techniques using \gls{tlstm}. \gls{tlstm} extends the \textquote{vanilla} \gls{lstm} cells by incorporating the elapsed time between consecutive events in order to adjust the memory state and is inspired by work from Baytas et al. \cite{baytas2017patient}.
\par 

Furthermore, the problem of next activity prediction is commonly modeled as a supervised multi-class classification problem. The distribution of activities in event logs are commonly skewed. Therefore, we additionally introduce cost-sensitive learning to address the inherent class-imbalances.

The main contributions of this work are summarized below:
\begin{itemize}
    \item We introduce a time-aware LSTM model for the tasks of predicting next activities and timestamps in \gls{pbpm}
    \item We tackle the problem of skewed class distributions via cost-sensitive learning
\end{itemize}

We evaluate the effectiveness of our proposed techniques by conducting experiments for the next activity and timestamp prediction on publicly available benchmark event logs commonly used for \gls{pbpm}.

\par
The remainder of the paper is structured as follows: Section~\ref{sec:rel_works} presents related work on DL-based next activity and timestamp prediction. 
Section~\ref{sec:back} introduces preliminaries and the concept of a LSTM.
Section~\ref{sec:meth} and \ref{sec:exp} describes the architecture of \gls{tlstm} and our experimental setup respectively.
Then, in Sections~\ref{sec:res} and \ref{sec:dis}, we present and discuss our results.
Section~\ref{sec:con} concludes our paper and discusses future research directions.

\section{Related Work} 
\label{sec:rel_works}
Inspired by the field of natural language processing (NLP), Evermann et al. \cite{evermann2017predicting} applied recurrent neural network-based and LSTM-basd DNN architectures for the next activity and next sequence of activity prediction in PBPM. They made use of word embeddings to encode activities of event log's process instances.
\par
Navarin et al. \cite{navarin2017lstm} used a \say{vanilla} LSTM-based DNN architecture for predicting the completion time of running process instances. They one-hot encoded the activity attributes, computed temporal control-flow attributes, and considered additional real-valued or categorical context attributes.
\par
Tax et al.\cite{tax2017predictive} proposed a multitask learning approach using \say{vanilla} LSTM cells for next activity and timestamp prediction respectively. Like in \cite{navarin2017lstm}, they one-hot encoded the activity and computed temporal control-flow features. However, the authors did not consider additional data attributes in their approach. This work acts as a baseline for a variety of other techniques such as \cite{weinzierl2020best}. 

Khan et al. \cite{khan2018memory} introduced memory augmented neural networks (MANNs) in \gls{pbpm}. MANNs reduce the number of trainable parameters. In general, the network's architecture consists of an externalized state memory and two \say{vanilla} LSTM cells manipulating the memory. One LSTM cell works as encoder and the other one as decoder. Concerning the predictive quality, their approach is comparable to the one presented in \cite{tax2017predictive}. 
\par
Camagro et al. \cite{camargo2019learning} extended the implementation of \cite{tax2017predictive} and fed the resource attribute into the DNN model. Additionally, instead of one-hot encoding, they applied embeddings, as proposed by Evermann et al.~\cite{evermann2017predicting}. 

Taymouri et al. \cite{taymouri2020predictive} introduced generative adversarial networks (GANs) for the next activity and timestamp prediction. The network's architecture comprises two \say{vanilla} LSTM cells. One for the generator and the other one for the discriminator.   
\par


To date, several studies have investigated DNN-based PBPM techniques. None of the related works proposes a DL-architecture that explicitly models the elapsed time between two successive events. We address this gap by adapting time-aware LSTM cells \cite{baytas2017patient}. 
Further, Mehdijev et al. \cite{Mehdiyev_Evermann_Fettke_2020} tackle the class imbalance problem in the context of the DNN-based prediction of next activities through a second neural network, namely radial basis function neural network, which generates semi-artificial data of the minority class in the pre-processing phase. In contrast, we adapt cost-sensitive learning to investigate the class-imbalance problem for DL-architectures comprising T-LSTM cells. 




\section{Background}
\label{sec:back}

\subsection{Preliminaries}

\begin{definition}[Event, Trace, Event Log]
\label{def:event_trace_log}
An event is a tuple $(c,a,ts)$ where $c$ is the case id, $a$ is the activity (label) and $ts$ is the timestamp. A trace is a non-empty sequence $\sigma = \langle e_{1}, \ldots, e_{\vert \sigma \vert} \rangle$ of events such that $\forall i, j \in \{1, \ldots, \vert \sigma \vert\}~ e_{i}.c = e_{j}.c$ and $e_{i}.ts \leq e_{j}.ts,$ for $1 \leq i < j \leq \vert \sigma \vert$. An event log $L$ is a set $\{\sigma_{1}, \ldots, \sigma_{\vert L \vert}\}$ of traces. A trace can also be considered as a sequence of vectors which contain derived control flow information or features. Formally, $\sigma=\left\langle\mathbf{x}^{(1)}, \mathbf{x}^{(2)}, \ldots, \mathbf{x}^{(\vert \sigma \vert)}\right\rangle$, where $\mathbf{x}^{(t)} \in \mathbb{R}^{{n} \times 1}$ is a vector, and the
superscript indicates the time-order upon which the events happened. $n$ is the number of features derived for each event. 
\end{definition}

\begin{definition}[Prefix and Label]
\label{def:prefix_label}
Given a trace $\sigma=\left\langle e_{1},\dots, e_{k}, \dots, e_{\vert \sigma \vert }\right\rangle$, a prefix of length $k$, that is a non-empty sequence, is defined as $f_{p}^{(k)}(\sigma)=\langle e_{1},\dots, e_{k}\rangle,$ with $0 < k < \vert \sigma_{c} \vert$. A next activity label for a prefix of length $k$ is defined as $f_{l,a}^{(k)}(\sigma)= e_{k+1}.a$, whereas a next timestamp label for a prefix of length $k$ is defined as $f_{l,ts}^{(k)}(\sigma)= e_{k+1}.ts$.     
The above definition also holds for an input trace representing a sequence of vectors. For example, the tuple of all possible prefixes, the tuple of all possible next activity labels and the tuple of all possible next timestamp labels for 
$\sigma=\langle\mathbf{x}^{(1)}, \mathbf{x}^{(2)}, \mathbf{x}^{(3)} 
\rangle$
are 
$\langle
\langle\mathbf{x}^{(1)}\rangle,
\langle\mathbf{x}^{(1)},\mathbf{x}^{(2)}\rangle
\rangle$,
$\langle e_2.a, e_3.a \rangle$,
and
$\langle e_2.ts,
e_3.ts
\rangle$.
\end{definition}

\subsection{Long Short-term Memory Cells}
\label{sec:lstm}

Most of the DNN architectures proposed for the next activity and timestamp prediction in PBPM \cite{weinzierl.2020} use \say{vanilla} LSTM cells \cite{hochreiter.1997}. LSTMs belong to the class of recurrent neural networks~\cite{lecun.2015} and are designed to handle temporal dependencies in sequential prediction problems~\cite{bengio.1994}.

Given a sequence of inputs $\sigma=\langle\mathbf{x}^{(1)}, \mathbf{x}^{(2)}, \mathbf{x}^{(3)}, ..., \mathbf{x}^{(k)}\rangle$, a LSTM computes sequences of outputs
$\langle\mathbf{h}^{(1)}, \mathbf{h}^{(2)}, \mathbf{h}^{(3)}, ..., \mathbf{h}^{(k)}\rangle$ via the following recurrent equations:

\vspace{-0.4cm}

\begin{align}
\label{eq:lstm}
    \mathbf{f}_{g}^{(t)}&=sigmoid(\mathbf{U}_{f} \mathbf{h}^{(t-1)}+\mathbf{W}_{f} \mathbf{x}^{(t)}+\mathbf{b}_{f})  & \text{(forget gate)}, \nonumber \\ 
    \mathbf{i}_{g}^{(t)}&= sigmoid(\mathbf{U}_{i} \mathbf{h}^{(t-1)}+\mathbf{W}_{i} \mathbf{x}^{(t)}+\mathbf{b}_{i}) & \text{(input gate)}, \nonumber \\
    \mathbf{\Tilde{c}}^{(t)}&= tanh(\mathbf{U}_{g} \mathbf{h}^{(t-1)}+\mathbf{W}_{g} \mathbf{x}^{(t)}+\mathbf{b}_{g}) & \text{(candidate memory)}, \nonumber \\
    \mathbf{c}^{(t)}&= \mathbf{f}_{g}^{(t)}  \circ \mathbf{c}^{(t-1)}+\mathbf{i}_{g}^{(t)} \circ \mathbf{\Tilde{c}}^{(t)} & \text{(current memory)}, \\ 
    \mathbf{o}_{g}^{(t)}&= sigmoid(\mathbf{U}_{o} \mathbf{h}^{(t-1)}+\mathbf{W}_{o} \mathbf{x}^{(t)}+\mathbf{b}_{o}) & \text{(output gate)}, \nonumber \\
    \mathbf{h}^{(t)}&= \mathbf{o}_{g}^{(t)} \circ tanh(\mathbf{c}^{(t)}) & \text{(current hidden state)}, \nonumber \\
    &\forall t \in \{1,2,\dots, k\}. \nonumber
    \end{align}

$\{\mathbf{U}_{f, i, g, o}, \mathbf{W}_{f, i, g, o}, \mathbf{b}_{f, i, g, o}\}$ are trainable parameters,  $\circ$ denotes the Hadamard product (element-wise product), $\mathbf{h}^{(t)}$ and $\mathbf{c}^{(t)}$ are the hidden state and cell memory of a LSTM cell. 
Additionally, a LSTM cell uses four gates to manage its states over time to avoid the problem of exploding/vanishing gradients in the case of longer sequences~\cite{bengio.1994}.
$\mathbf{f}_{g}^{(t)}$ (forget gate) determines how much of the previous memory is kept, $\mathbf{i}_{g}^{(t)}$ (input gate) controls the amount new information is stored into memory, $\mathbf{\Tilde{c}^{(t)}}$ (candidate memory) defines how much information is stored into memory and $\mathbf{o}_{g}^{(t)}$ (output gate) determines how much information is read out of the memory. The hidden state $\mathbf{h}^{(t)}$ is commonly forwarded to a successive layer.
\section{Methodology}
\label{sec:meth}

\subsection{Time-Aware Long Short-term Memory Cells}
\label{sec:tlstm}
\begin{figure}[htb]
\minipage{0.5\textwidth}
\centerline{\includegraphics[width=0.9\textwidth]{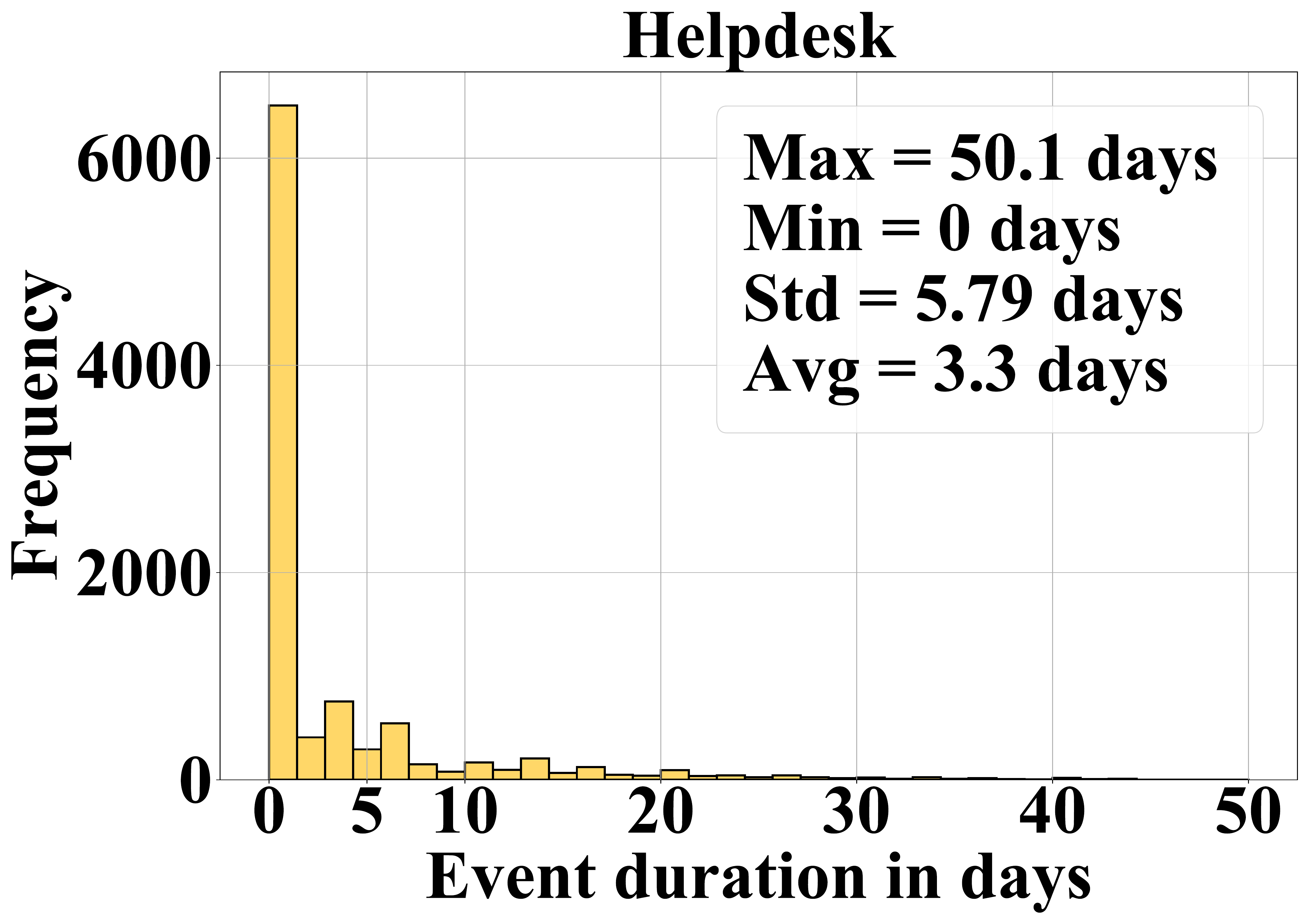}}
\endminipage
\minipage{0.5\textwidth}
\centerline{\includegraphics[width=0.9\textwidth]{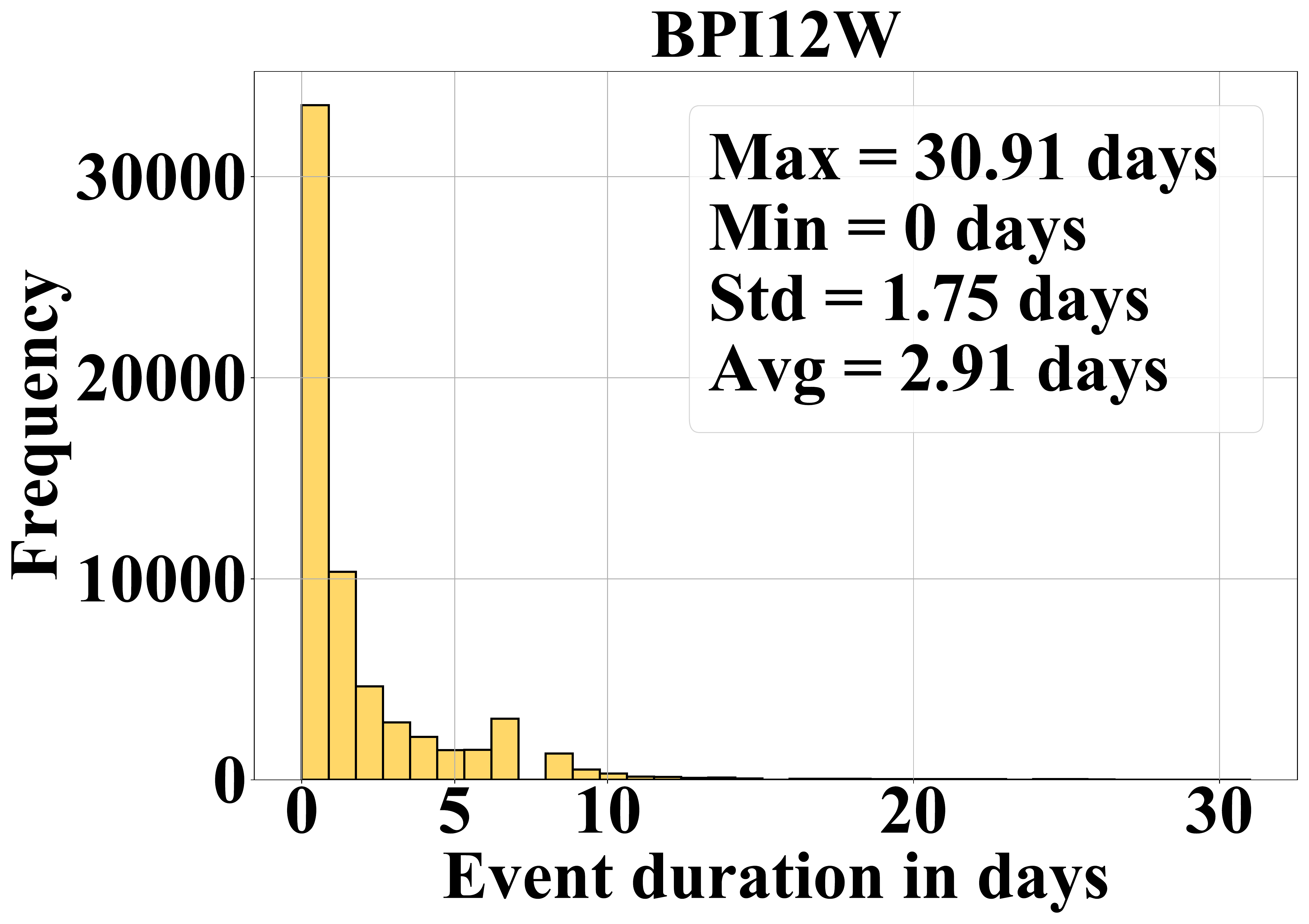}}
\endminipage\\
\caption[]{Event duration distribution for the complete Helpdesk and BPI12W datasets. It can be observed that the  majority of the events are completed within one day. However, there are many events with longer duration. Note that we input the elapsed time between events ($\mathbf{\Delta}^{t}$) in seconds for T-LSTM.}
\label{fig:tmests}
\end{figure}

\say{Vanilla} LSTM cells, as described in Section \ref{sec:lstm}, assume a uniform distribution of the elapsed time between events $(\mathbf{\Delta}^{(t)} \coloneqq {x_{ts}}^{(t)}-{x_{ts}}^{(t-1)})$. This assumption does not hold for most event logs analyzed in PBPM though (see Fig. \ref{fig:tmests}). The elapsed time between consecutive events might have an impact on the next activity and timestamp prediction. Hence, a LSTM cell should be able to take irregular elapsed times into account when processing event logs.
\par
\begin{figure}[htpb]
\centering
\includegraphics[width=0.9\textwidth]{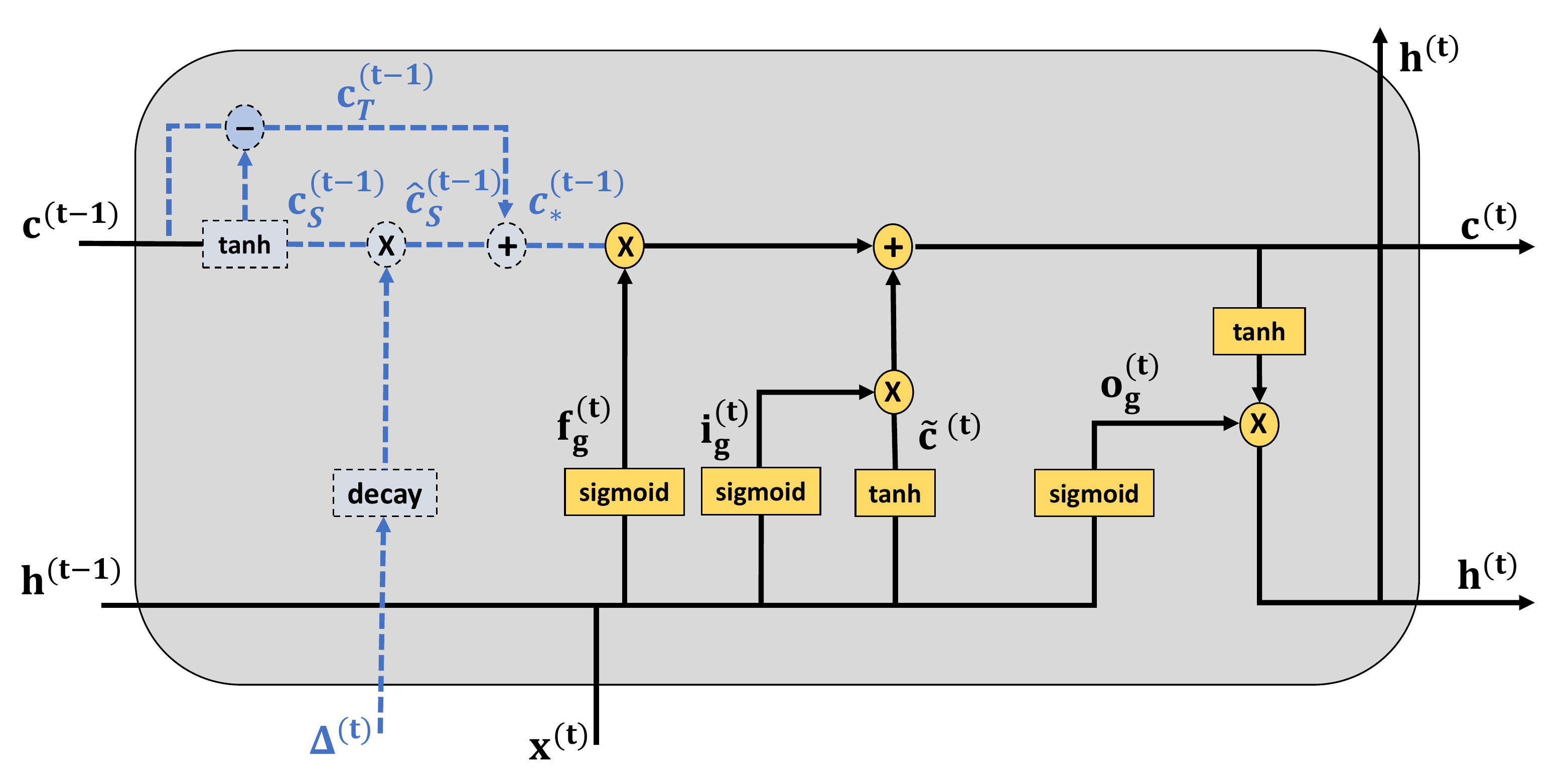}
\caption[]{Illustration of a T-LSTM cell with its computational components at time step $t$. The dashed and blue components indicate the extensions to the \say{vanilla} LSTM cell. The previous cell memory $\mathbf{c}_{S}^{(t-1)}$ is adjusted to $\mathbf{c}_{*}^{(t-1)}$ (see Eq. \eqref{eq:tlstm}) and is then processed together with $\mathbf{h}^{(t-1)}$ and $\mathbf{x}^{(t)}$ via the LSTM computations, as formalized in Eq. \eqref{eq:lstm}.} 
\label{fig:tlstm}
\end{figure}

Time-aware long short-term memory (T-LSTM) cells are an extension of the LSTM. Fig. \ref{fig:tlstm} depicts the T-LSTM cell and highlights its differences with regard to the LSTM cell.

The main idea behind T-LSTM is to perform a subspace decomposition of the previous cell memory $\mathbf{c}^{(t-1)}$. First, a short term memory component $\mathbf{c}_{S}^{(t-1)}$ is extracted via a network. Next, the short term memory is discounted via a decay function of the elapsed time and yields $\mathbf{\hat{c}}_{s}^{(t-1)}$. Then, the long term memory $(\mathbf{c}_{T}^{(t-1)}=\mathbf{c}^{(t-1)} - \mathbf{c}_{S}^{(t-1)})$ is calculated. Finally, the previous cell memory is adjusted $\mathbf{c}_{*}^{(t-1)}=\mathbf{c}_{T}^{(t-1)}+\mathbf{\hat{c}}_{s}^{(t-1)})$. The adjusted previous memory $\mathbf{c}_{*}^{(t-1)}$ is then, together with  $\mathbf{h}^{(t-1)}$ and $\mathbf{x}^{(t)}$, further processed as in LSTM cells by substituting $\mathbf{c}^{(t-1)}$ with $\mathbf{c}_{*}^{(t-1)}$ in Eq. \eqref{eq:lstm}. The following equations summarize the T-LSTM specific computations for the subspace decomposition and adjustment of the previous memory.    

\begin{align}
\label{eq:tlstm}
    \mathbf{c}_{S}^{(t-1)}&=tanh(\mathbf{W}_{d} \mathbf{c}^{(t-1)}+\mathbf{b}_{d})  &  \text{(short term memory)}, \nonumber \\
    \mathbf{\hat{c}}_{s}^{(t-1)}&=\mathbf{c}_{S}^{(t-1)}*decay(\mathbf{\Delta}^{(t)})  &  \text{(discounted short term memory)}, \nonumber \\
    \mathbf{c}_{T}^{(t-1)}&=\mathbf{c}^{(t-1)}-\mathbf{c}_{S}^{(t-1)}  &  \text{(long term memory)},  \\
    \mathbf{c}_{*}^{(t-1)}&=\mathbf{c}_{T}^{(t-1)}+\mathbf{\hat{c}}_{s}^{(t-1)}  &  \text{(adjusted previous memory)}, \nonumber \\
    &... & \text{(LSTM computations as in Eq. \eqref{eq:lstm})},  \nonumber \\
    &\forall t \in \{1,2,\dots, k\}.\nonumber
\end{align}

Note, that we only add $\{\mathbf{W}_{d}, \mathbf{b}_{d}\}$ as trainable parameters compared to the LSTM cell. As recommended in Baytas et al. \cite{baytas2017patient}, we chose $decay(\mathbf{\Delta}^{(t)}) = 1/log(e+\mathbf{\Delta}^{(t)})$ since we input the elapsed times in seconds and therefore have large values for $\mathbf{\Delta}^{t}$. Any other monotonic decreasing function and scale for $\mathbf{\Delta}^{t}$ would be valid as well, but our initial choice proved to be effective. The intuition behind the subspace decomposition is that the short term memory should be discounted if the elapsed time is very large, while the long term memory should be maintained in the adjusted previous cell memory $\mathbf{c}_{*}^{(t-1)}$. Similar as for LSTMs, the hidden state $\mathbf{h}^{(t)}$ is forwarded to successive layer for further processing. Hence, it is straightforward to substitute LSTM with T-LSTM cells in a given DNN architecture.   

\subsection{Network Architecture}
\begin{figure}[hbt!]
\centering
\includegraphics[scale = 0.5]{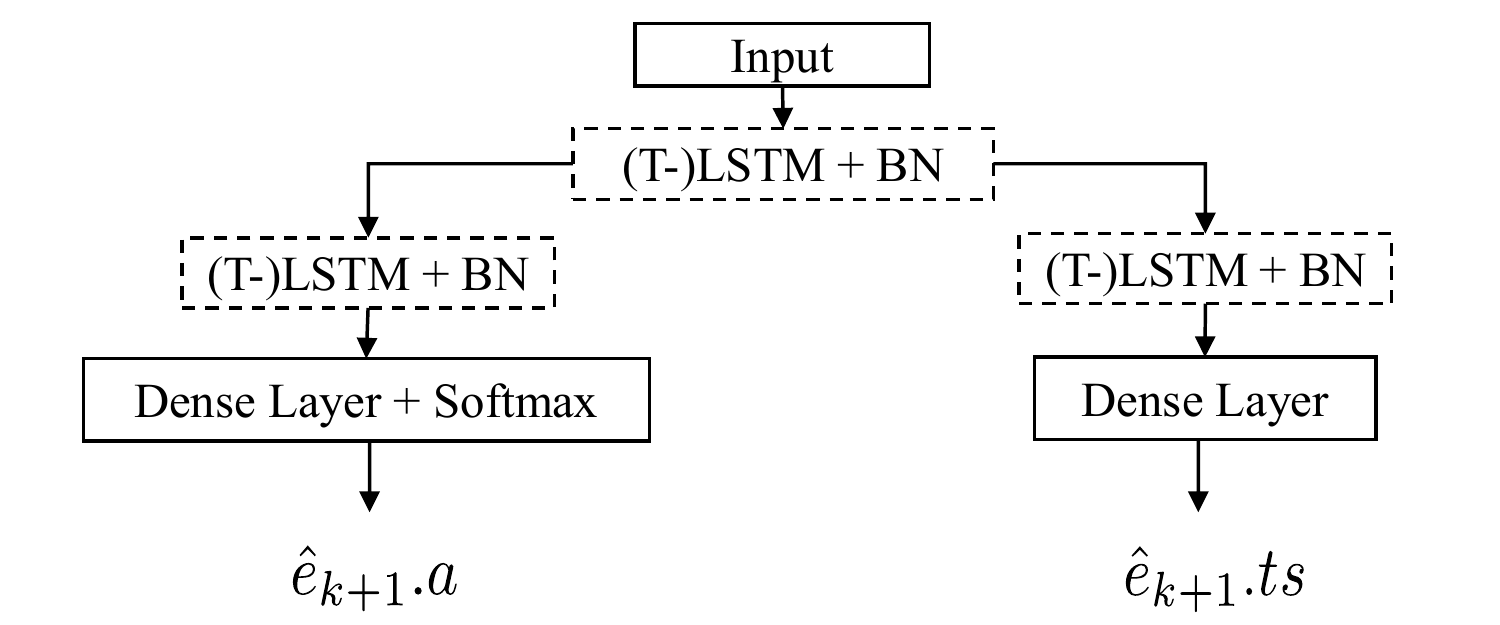}
\caption{Network architecture for this work based on the multitask learning approach proposed by Tax et al. \cite{tax2017predictive}. The dashed components are either LSTM or T-LSTM layers. The input is of the network is a sequence of vectors representing a prefix $\langle e_{1},\dots, e_{k}\rangle$ as in Tax et al. \cite{tax2017predictive}. For the baseline architecture we applied one-hot encoding and LSTM layers as in \cite{tax2017predictive}. The outputs of the model are the predicted next activity ($\hat{e}_{k+1}.a$) and timestamp ($\hat{e}_{k+1}.ts$). Each of the LSTM layers is followed by a batch normalization layer (BN) to speed up training, as used in Tax et al. \cite{tax2017predictive}.}
\label{fig:arch}
\end{figure}

We adapted the multitask architecture proposed by Tax et al. \cite{tax2017predictive} as a baseline (see Fig. \ref{fig:arch}). The predicted next activity 
$\hat{e}_{k+1}.a$ is the output of a softmax activation after the last dense layer, where the output dimension is equal to the number of unique activity labels. $\hat{e}_{k+1}.a$ is evaluated against the one-hot encoded ground truth label $e_{k+1}.a$ by using the Cross-Entropy (CE) loss. 
The predicted next timestamp
$\hat{e}_{k+1}.ts$ is a scalar output of a dense layer. We do not apply any additional activation after the time specific dense layer to be consistent with the implementation\footnote{\url{https://github.com/verenich/ProcessSequencePrediction}} of Tax et al. \cite{tax2017predictive}. 
$\hat{e}_{k+1}.ts$ is compared with the ground truth timestamp $e_{k+1}.ts$ using the Mean Absolute Error (MAE). 
The total loss is the sum of both losses, as implemented in Tax et al. \cite{tax2017predictive}. Further, they applied one-hot encoding for the activities and compute time-related control-flow features, which we also used in our experiments. We refer to the baseline architecture as \say{\textbf{Tax}}. We performed an ablation study and made three modifications to the baseline DNN architecture:
\begin{itemize}
    \item We weighted the CE loss function based on the distribution of activity labels in the training set. Hence, the classification of under-represented event classes had larger influence during training. We refer to this model as \say{\textbf{Tax+CS}}.
    \item We replaced all LSTM layers with T-LSTM layers and refer to this model as \say{\textbf{Tax+T-LSTM}}.
    \item We added cost-sensitive learning and replaced all LSTM layers with T-LSTM layers. We call this model \say{\textbf{Tax+CS+T-LSTM}}
\end{itemize}

\section{Experiments}
\label{sec:exp}

\subsection{Datasets}

\begin{figure}[htb]
\minipage{0.5\textwidth}
\centerline{\includegraphics[width=0.9\textwidth]{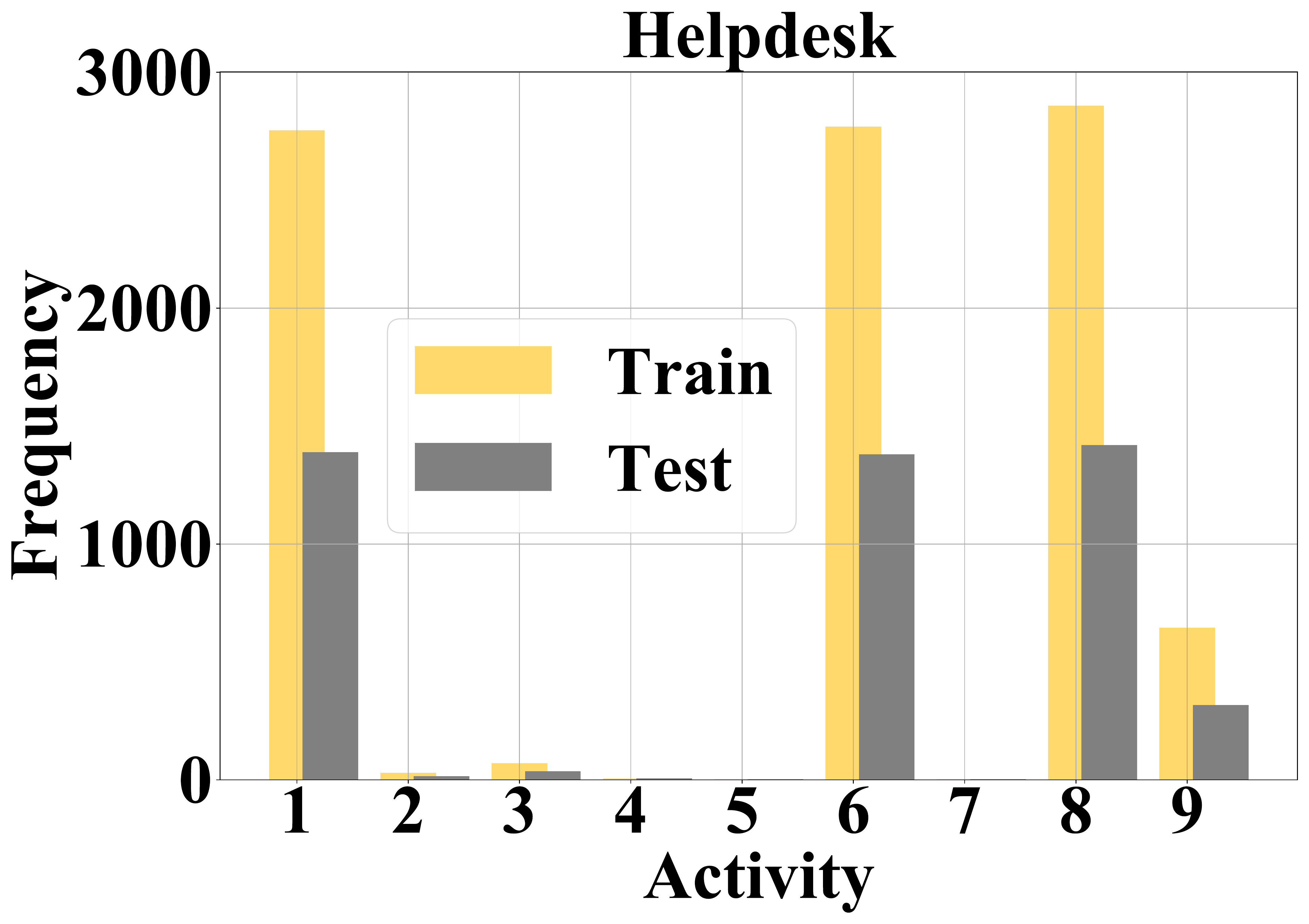}}
\endminipage
\minipage{0.5\textwidth}
\centerline{\includegraphics[width=0.9\textwidth]{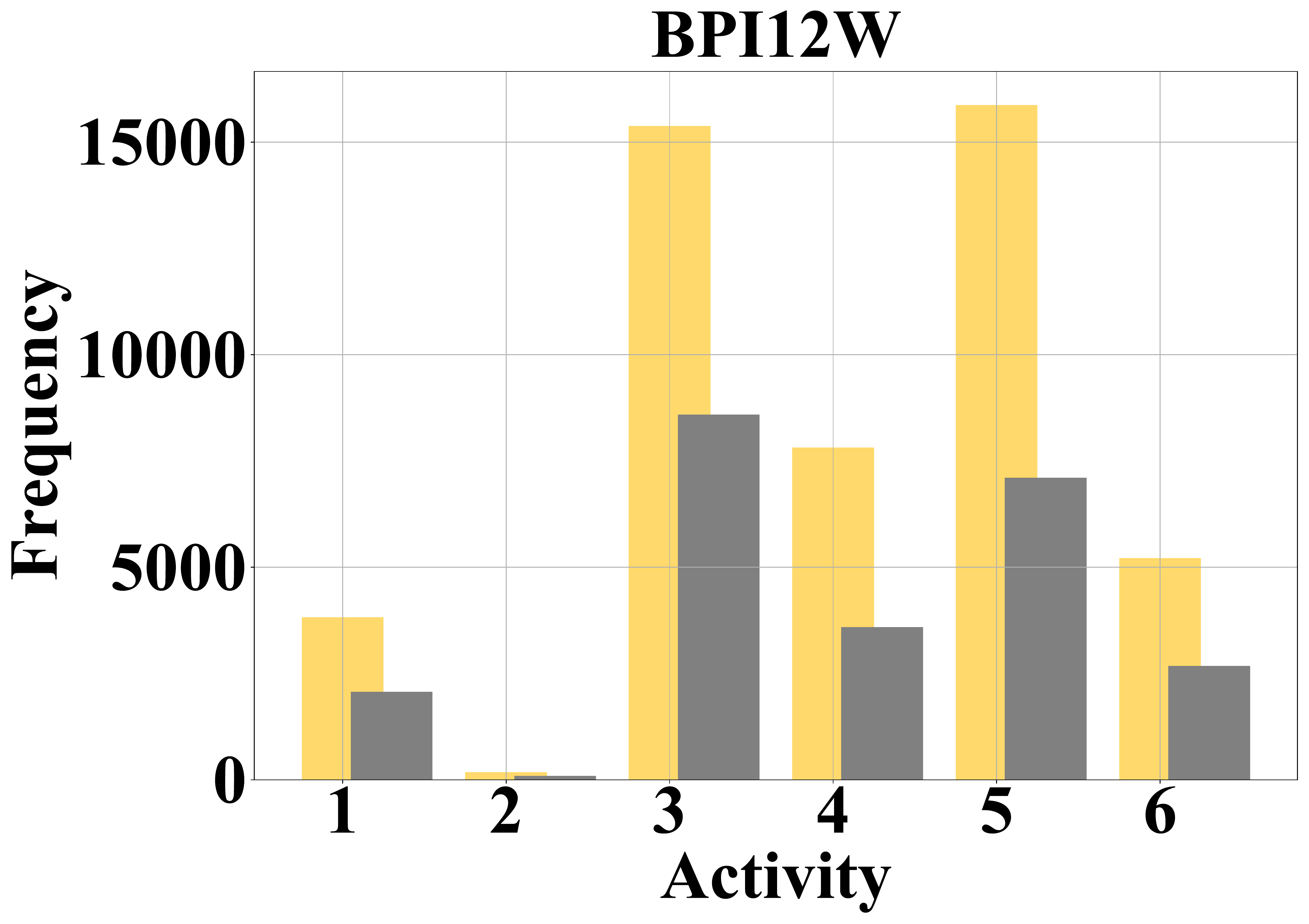}}
\endminipage\\

\caption{Activity distribution in training and test set for Helpdesk and BPI12W datasets. It is evident that the distributions of the activity labels are skewed.}
\label{fig:cls}
\end{figure}

We performed our experiments on the same publicly available datasets as Tax et al. \cite{tax2017predictive} to validate the effectiveness of our proposed techniques. Fig. \ref{fig:cls} shows the distribution of the activities (labels) for the different datasets. It is evident that the distributions of activities are skewed for both event logs. Table \ref{tab:data_stats} presents descriptive statistics of the datasets used in this work.

 \noindent \textbf{Helpdesk\footnote{\url{https://doi.org/10.4121/uuid:0c60edf1-6f83-4e75-9367-4c63b3e9d5bb}}:} This event log originates from a ticket management process of an Italian software company.
 
 \noindent \textbf{BPI'12 W Subprocess\footnote{\url{https://doi.org/10.4121/uuid:3926db30-f712-4394-aebc-75976070e91f}} (BPI12W):} The Business Process Intelligence (BPI) 2012 challenge provided this event log from a German financial institution. The data come from a loan application process. The ‘W' indicates state of the work item for the application.
\begin{table}[htb]
  \begin{center}
    \begin{tabular}{|l|c|c|r|} 
    \hline
      \textbf{Characteristic } & \textbf{Helpdesk} & \textbf{BPI12W}\\
      \hline
      \hline
      Number of instances & 3,804 & 9,658\\
      \hline
      Case variants &  154& 2,263\\
      \hline
      Unique activities &9 & 6\\
      \hline
      Events  &  13,710 & 72,413\\
      \hline
      Max \# events per case & 14 & 74\\
      \hline
     Min \# events per case & 1 & 1\\
     \hline
      Avg \# events per case &  3.604 &  7.497\\
     \hline
     \end{tabular}
        \caption{Descriptive statistics of the datasets used in this study.}
         \label{tab:data_stats}
  \end{center}
\end{table}

\subsection{Preprocessing}

We used the cleaned and prepared datasets as in Tax et al.~\cite{tax2017predictive}. The datasets can be found on the corresponding GitHub repository\footnote{\url{https://github.com/verenich/ProcessSequencePrediction/tree/master/data}}. The preprocessing steps include splitting the data into training and test set, calculating time divisors, and ASCII encoding activities and sequence generation. Datasets were split into $2/3$rd and $1/3$rd for training and testing respectively and preserve the temporal order of cases. We additionally used the last $20\%$ of the training data as a validation set in order to tune the hyperparameters.
We adapted the sequence and feature generation methods by Tax et al. \cite{tax2017predictive}. The features include the activity of the event, position of the event in the case, time since the last event, time from the starting event of the case, time from midnight, and day of the week. We create one-hot encoded versions of the ground truth labels ${e}_{k+1}.a$ for the next activity prediction in order to compare them with the predicted next activity labels $\hat{e}_{k+1}.a$.

\subsection{Training Setup}
For hyperparameter tuning, we performed a grid search on the training set and chose the model with the lowest validation loss. The validation loss is the sum of activity-related validation loss and time-related validation loss. The number of LSTM or T-LSTM units were set to $64$ or $100$. For the dropout rate (for both dense layers), we tried the values $0.0$ and $0.2$. We choose Nadam as an optimization algorithm, as used in \cite{tax2017predictive}. Nesterov accelerated gradient (NAG) calculates the step using the ‘lookahead’ algorithm, which approximates the next parameters. Adam optimizer estimates learning rates based on initial moments of the gradients. Nadam is a combination of both and is robust in noisy datasets.
Furthermore, we tested a range of different learning rates $\{0.0001, 0.0002, 0.001, 0.002, 0.01\}$ since this is known to have a large impact on LSTMs \cite{searchspace_lstm}.
We trained each model for $150$ epochs, with a batch size of $64$ and apply early stopping with patience $25$ for regularization. 

\subsection{Evaluation}
We applied the same evaluation metrics as in \cite{tax2017predictive}. We used the \emph{Accuracy} metric to evaluate the next activity prediction. For the next timestamp prediction, we used the \emph{Mean Absolute Error (MAE)} to evaluate our models.

\subsection{Implementation}
We conducted all experiments on a workstation with 24 CPU cores, 748 GB RAM and a singe GPU NVIDEA QUADRO RTX6000. We implemented the experiments in Python 3.7. We used the DL framework TensorFlow 2.1\footnote{\url{https://www.tensorflow.org}}. The source code is available on GitHub\footnote{\url{https://github.com/annguy/time-aware-pbpm}}.

\section{Results}
\label{sec:res}
\subsection{Next Activity Prediction} Table \ref{tab:Acc} shows the results for the next activity prediction in terms of Accuracy.
For Helpdesk and BPI12W, the approach \mbox{Tax+CS+T-LSTM} achieved the highest Accuracy (0.724 and 0.778) among all approaches. The approach's improvement compared to the baseline is 0.012 and 0.018.
While the two approaches, \mbox{Tax+CS} and \mbox{Tax+T-LSTM}, outperformed the baseline for Helpdesk, these approaches achieved a lower Accuracy for BPI12W than the baseline.   

\begin{table}[h!]
  \begin{center}
    \begin{tabular}{|l|c|c|r|} 
    \hline
      \textbf{Approach} & \textbf{Helpdesk} & \textbf{BPI12W}\\
      \hline
      \hline
      \bf{Tax (baseline)} &0.712 & 0.760\\
      \hline
      \bf{Tax+CS}&0.713 & 0.757 \\
      \hline
      \bf{Tax+T-LSTM}&0.718 & 0.693 \\
      \hline
      \bf{Tax+CS+T-LSTM} & \textbf{0.724} & \textbf{0.778} \\
      \hline
    \end{tabular}
        \caption{Results for the next activity prediction in terms of Accuracy. The best result for each dataset is highlighted (larger is better).}
         \label{tab:Acc}
  \end{center}
\end{table}

\subsection{Next Timestamp Prediction} Table~\ref{tab:MAE} shows the results for the next timestamp prediction task in terms of MAE in days. 
All approaches with a T-LSTM cell, clearly outperformed the baseline for both event logs.
Thereby, the approach \mbox{Tax+CS} achieved the lowest MAE of 2.87 days and 0.88 days for Helpdesk and BPI12W respectively. Compared to the baseline, this approach reduced the MAE by 0.88 days (Helpdesk) and 0.68 days (BPI12W).  
The other two approaches, \mbox{Tax+T-LSTM} and \mbox{Tax+CS+T-LSTM}, achieved a slightly worse MAE values compared to \mbox{Tax+CS} for both event logs. 
It is worth noticing that for Helpdesk \mbox{Tax+CS+T-LSTM} and for BPI12W \mbox{Tax+T-LSTM} yielded the second best results with MAE close to \mbox{Tax+CS}. 

   \begin{table}[h!]
  \begin{center}
    \begin{tabular}{|l|c|c|r|} 
    \hline
      \textbf{Approach} & \textbf{Helpdesk} & \textbf{BPI12W}\\
      \hline
      \hline
      \bf{Tax (baseline)} & 3.75 & 1.56\\
      \hline
      \bf{Tax+CS} & \textbf{2.87} & \textbf{0.88}\\
      \hline
      \bf{Tax+T-LSTM} & 3.01&0.88\\
      \hline
      \bf{Tax+CS+T-LSTM} & 2.94&0.90\\
      \hline
    \end{tabular}
        \caption{Results for next step time prediction in terms of MAE in days. The best result for each dataset is highlighted (lower is better).}
         \label{tab:MAE}
  \end{center}
\end{table}

\section{Discussion}
\label{sec:dis}
In this paper, we argued that the elapsed time between consecutive events carries valuable information on human behavior in running business processes.
Therefore, we introduced T-LSTM cells for PBPM which inherently model the elapsed time between consecutive events. Further, we introduced of cost-sensitive learning to better cope with the problem of imbalanced data.  

The obtained results indicate that the elapsed time between consecutive events is informative and that a DNN architecture relying on T-LSTM cells cab yield more accurate models for PBPM. 
Especially, with the approach \\ \mbox{Tax+CS+T-LSTM}, we could outperform the baseline (Tax) for both datasets (i.e., Helpdesk and BPI12W) and both prediction tasks (i.e., next activity prediction and next timestamp prediction).
Thereby, we could observe that cost-sensitive learning plays a crucial role for the predictive quality of a DNN architecture using T-LSTM cells instead of \say{vanilla} LSTM cells. 
Interestingly, the effectiveness of the introduced techniques is more evident for the next timestamp prediction compared to the next activity prediction
\par
Even though our presented results on DNN architectures using T-LSTMs seem promising, there are a few limitations to our work.
First, we need to verify our findings by performing experiments on more datasets. 
Second, a better hyperparameter tuning approach like Bayesian optimization \cite{optuna_2019} could be applied for all configurations to get a better estimate of their effectiveness. Further, several runs with random initialization should be performed to estimate the stability of the models.  
\section{Conclusion and Future Work} 
\label{sec:con}
We propose T-LSTM as an alternative to the commonly used \say{vanilla} LSTM cell to better exploit information on the elapsed time between consecutive events. Furthermore, we introduced the concept of cost-sensitive learning to account for the common class-imbalance in event log data. Our results indicate the effectiveness of the introduced techniques for the next activity and timestamp prediction. This suggests that integrating specific mechanisms into neural network layers to incorporate event log specific characteristics might be an interesting direction for future research. Here, we mainly demonstrated the benefit of replacing a normal LSTM with a time-aware LSTM cell for a given baseline approach \cite{tax2017predictive}. 
\par 
An avenue for future research is to investigate if T-LSTM cells might also improve other LSTM-based PBPM approaches such as Camargo et al. \cite{camargo2019learning} involving resource attributes or Taymouri et al. \cite{taymouri2020predictive} generating fake event logs. 
Another direction for future research is to further customize an LSTM cell in terms specifically for PBPM. For example, a process-aware LSTM cell could not only deal with time information but also with resource information. 

\bibliographystyle{splncs04.bst}
\bibliography{mad}

\end{document}